\documentclass{article}

\usepackage{arxiv}

\usepackage[utf8]{inputenc} 
\usepackage[T1]{fontenc}    
\usepackage{url}            
\usepackage{booktabs}       
\usepackage{amsfonts}       
\usepackage{nicefrac}       
\usepackage{microtype}      
\usepackage{lipsum}
\usepackage{xcolor}
\usepackage{graphicx}
\usepackage{subfigure}
\usepackage{amsmath}
\usepackage[hidelinks]{hyperref}

\usepackage{color,soul}

\title{Stress Field Prediction in Cantilevered Structures Using Convolutional Neural Networks}

\author{
  Zhenguo Nie \\
  Department of Mechanical Engineering\\
  Carnegie Mellon University\\
  Pittsburgh, PA 15213\\
  \texttt{zhenguon@andrew.cmu.edu} \\
   \And
  Haoliang Jiang \\
  Department of Mechanical Engineering\\
  Carnegie Mellon University\\
  Pittsburgh, PA 15213\\
  \texttt{haolianj@andrew.cmu.edu} \\
  \And
  Levent Burak Kara\thanks{Address all correspondence to this author.} \\
  Department of Mechanical Engineering\\
  Carnegie Mellon University\\
  Pittsburgh, PA 15213\\
  \texttt{lkara@cmu.edu} \\
}

\begin{document}
\maketitle

\begin{abstract}
\textit{The demand for fast and accurate structural analysis is becoming increasingly more prevalent with the advance of generative design and topology optimization technologies. As one step toward accelerating structural analysis, this work explores a deep learning based approach for predicting the stress fields in 2D linear elastic cantilevered structures subjected to external static loads at its free end using convolutional neural networks (CNN). Two different architectures are implemented that take as input the structure geometry, external loads, and displacement boundary conditions, and output the predicted von Mises stress field. The first is a single input channel network called SCSNet as the baseline architecture, and the second is the multi-channel input network called StressNet. Accuracy analysis shows that StressNet results in significantly lower prediction errors than SCSNet on three loss functions, with a mean relative error of 2.04\% for testing. These results suggest that deep learning models may offer a promising alternative to classical methods in structural design and topology optimization. Code and dataset are available at \url{https://github.com/zhenguonie/stress_net}}

\end{abstract}


\section{INTRODUCTION}

While computational stress analysis is fundamentally critical in design and engineering, advances in automatic generative design systems impose higher demands on analysis speed without compromising accuracy. In this work, we explore the potential of data-driven stress analysis, where conventional run-time analysis is replaced by a machine learning system that can generate solutions instantaneously. Our approach is inspired by the great and increasing success of data-driven approaches that model physical phenomena and use the acquired knowledge to make predictions for unseen problems. Recent advances include data-driven approaches demonstrated in fluid dynamics \cite{2,3,4,5}, design and topology optimization \cite{6,7,8,9}, molecular dynamics simulation \cite{14,15,16,17}, and others \cite{10,18,19,21,point_cloud}. 

In computational solid mechanics, there have been early attempts to use neural regression for finite element analysis (FEA) model updating \cite{FEA_modelUpdate1, FEA_modelUpdate2}. More recently, deep learning has been gaining interest in the solution of traditional mechanics problems. Javadi et al. \cite{NN_FAE} use a simple neural network in FEA as an alternative to the conventional constitutive material model. However, the geometry is abstracted and simplified into a feature vector, making the approach difficult to generalize to complex cases. Deep learning has also been integrated into FEA to optimize the numerical quadrature in the element stiffness matrix on a per-element basis \cite{23}. Compared to the conventional numerical approach, this approach accelerates the calculation of the element stiffness matrix. 

The use of convolutional networks, which uses learned filters to understand image-based representations, has caught the attention of the academic community these years. Spruegel et al. \cite{NN_plausibilityCheck} use a convolutional network as a classifier to check the plausibility of FE simulations. In an inspiring work, Liang et al. \cite{surrogate_FEA} develop a three-module convolutional network for aortic wall stress prediction to accelerate the patient-specific FEA. The network takes as input the tube-shaped geometry and outputs the stress field. The description of geometry, boundary conditions and loads, is tailored for aorta models. We build upon these advances to demonstrate the potential of data-driven techniques for stress field predictions in a moderately more general case. 

In this work, we present an end-to-end deep learning based approach to predict the stress field in 2D linear elastic deformations. Two different architectures are explored that take as input the geometry, loads, and boundary conditions, and output the predicted von Mises stress field. The first architecture is a single-channel stress prediction neural network (SCSNet) where the loads are augmented with the feature representation (FR). The second is a multi-channel stress prediction neural network (StressNet) with five separate channels of input including the geometry, load and boundary conditions represented by images of identical sizes. In principle, StressNet can be extended to a generalized model for any arbitrary 2D conditions once given the corresponding training data, due to its on-limits multiple input channels.

This image-based problem representation allows the encoding of arbitrary 2D structures (within the prescribed domain resolution), boundary conditions and external forces. As one step toward assessing the feasibility of such an approach, we focus on cantilevered structures with loads applied to the free end of the structure. A dataset involving 120,960 problems with variations in geometry and loads is generated to train and evaluate the networks. In our tests, StressNet with a deep architecture has a significantly higher accuracy over SCSNet, with a mean relative stress error of 2.04\%. An inbetween network that combines elements of SCSNet and StressNet is also studied to assess the impact of single versus multi-channel input representation.




\section{BACKGROUND and RELATED WORK}
\smallskip
\noindent
\textit{Finite element analysis for stress computation.} Stress analysis of a given structure requires the solution of related partial differential equations. Finite element analysis (FEA) is the conventional approach to solve this problem. It simplifies the structure by breaking it down into a large number of finite elements and computes the coupled mechanical deformations and stresses based on the boundary and load conditions by building up an algebraic equation:
\begin{equation}
Ku=F 
\label{Eq:KQ}
\end{equation}
where $K$ is the global stiffness matrix, $F$ is the vector describing the applied external nodal forces, $u$ denotes the nodal displacement vector. To compute displacements, the global stiffness matrix has to be first assembled.

We assume linear isotropic materials and small deformations in a 2D field. The elemental stiffness matrix $\boldsymbol{K}_e$ can be computed as follows:

\begin{equation}
\boldsymbol{K}_e = A_e \boldsymbol{B}_e^T \boldsymbol{C}_e \boldsymbol{B}_e,
\label{Eq:elementStiffness}
\end{equation}

\noindent where $A_e$ is the area of the element, $\boldsymbol{B}_e$ is the strain-displacement matrix that depends only on the element's rest shape and $\boldsymbol{C}_e$ is the elasticity tensor constructed using Young's modulus and Poisson's ratio of the base material.
Given a mesh $\mathbb{V}$ with $m$ elements, one can assemble the global stiffness matrix $\boldsymbol{K}$ in order to determine the displacements $\boldsymbol{u}$ from \eqref{Eq:KQ} through a proper application of the zero displacement boundary conditions and the external nodal forces. Then, the stress-displacement relationship can be written as:

\begin{equation}
\boldsymbol{\sigma} = \boldsymbol{C_g} \boldsymbol{B} \boldsymbol{u}, 
\label{Eq:stressFormula}
\end{equation}

\noindent where $\boldsymbol{\sigma} \in \mathbb{R}^{4m}$ captures the unique four elements of the elemental stress tensor for planar stress calculations
and $\boldsymbol{B}$ is the assembly of $\boldsymbol{B}_e$ matrices. Block-diagonal matrix $\boldsymbol{C}_g \in \mathbb{R}^{4m \times 4m}$ is constructed with elemental elasticity tensors $\boldsymbol{C}$ on the diagonal.
For each element, $\boldsymbol{C}_e$ can be computed analogous to $\boldsymbol{K}_e$.
While applicable to different element types, we use linear quad elements making $\boldsymbol{K} \in \mathbb{R}^{2n \times 2n}$, $\boldsymbol{u} \in \mathbb{R}^{2n}$, $\boldsymbol{f} \in \mathbb{R}^{2n}$ and $\boldsymbol{B} \in \mathbb{R}^{4m \times 2n}$ for a planar mesh with quad elements having $n$ nodes.

\noindent Then, the von Mises Stress of each element is computed using 2-D von Mises Stress form:
\begin{equation}
\sigma_{vm} = \sqrt{\sigma{x}^2 + \sigma_{y}^2 - \sigma_{x}\sigma_{y}+3\tau_{xy}^2}
\end{equation}
where $\sigma_{vm}$ is von Mises Stress, $\sigma_{x}$ and $\sigma_{y}$ are the stress components in the $x$ and $y$ directions respectively, and $\tau_{xy}$ is the shear stress. 

The final stress distribution could be obtained after the stresses of all elements are computed. Several factors impact FEA's complexity including the number of elements and the degree of the elements, as they give rise to large stiffness matrices that need to be assembled from elemental stiffness matrices \cite{surrogate_FEA}. This leaves an opportunity for researchers to seek faster methods for inner loop simulations for structure design and optimization. This work explores the use of deep learning toward this goal.

\medskip
\noindent
\textit{Convolutional neural networks and ResNet.} Convolutional neural networks (CNN) are designed to process data that comes in the form of multiple arrays, for example, a color image composed of three 2D arrays containing pixel intensities in the three-color channels \cite{1}. CNN consists primarily of two core structures: a convolutional layer and a pooling layer. In a convolutional layer, a filter bank with a set of weights slides over the input image to produce a feature map. After a convolutional layer, the result of the weighted sum is usually passed through a non-linearity such as a ReLU function: $f(x)={\rm max}(0,x)$, and then passed on to the pooling layer. The role of the pooling layer is to merge semantically similar features into one.

\begin{figure}[ht]
	\centering
	\includegraphics[width=0.6\linewidth]{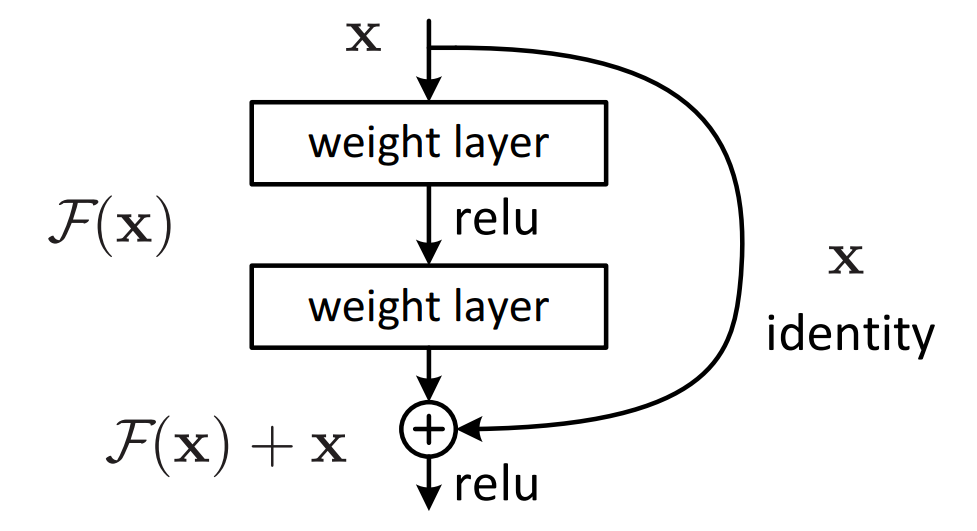}
	\caption{ResNet: a building block with a shortcut connection \cite{27}}
	\label{fig: resnet}
\end{figure}

Inspired by the philosophy of VGG net \cite{simonyan2014very}, He et al. \cite{27} propose a 152-layer Residual Network (ResNet) that won the first place in ILSVRC-2015 with an error rate of 3.6 \%. As shown in Figure \ref{fig: resnet}, the defining feature of ResNet is the shortcut connection added to each pair of $3 \times 3$\ filters in the residual version. The shortcut connection simply performs identity mapping, and its output is added to the output of the stacked layers. To the extreme, if an identity mapping was optimal, it would be easier to push the residual to $0$ than to fit an identity mapping by the stacked layers. It means ResNet can dynamically select the layer depth for the desired underlying mapping.

\medskip
\noindent
\textit{Deep learning in computational mechanics.} The Artificial Neural Networks (ANN) with the multilayer perceptron (MLP) has been applied in computational mechanics for several years, for structural analysis \cite{jenkins1995neural,waszczyszyn2001neural,goh1995multivariate}, materials failure and damage \cite{abendroth2003determination,wu1992use,zang2001structural,tsou1994structural}, regression of the material constitutive properties \cite{abendroth2003determination,huber1999determination,zhang2003artificial,settgast2019constitutive,liu2019estimation}, computational mechanics enhancement \cite{23,NN_FAE}, topological design and optimization \cite{yildiz2003integrated,papadrakakis2002reliability}. However, due to the intrinsic limitation of MLP, the geometry is usually abstracted and simplified into a feature vector that loses  the spatial position relationship, making the approach difficult to generalize to complex cases.

Spruegel et al. \cite{NN_plausibilityCheck} use a convolutional network as a classifier to check the plausibility FE simulations. Similarly, Sosnovik et al. \cite{sosnovik2017neural} propose a convolutional encoder-decoder architecture to accelerate topology optimization methods. Khadilkar et al. \cite{khadilkar2019deep} use CNN to predict the stress field for the bottom-up SLA 3D printing process. In an inspiring work, Liang et al. \cite{surrogate_FEA} develop a three-module convolutional network for aortic wall stress prediction to accelerate the patient-specific FEA. The network takes as input the tube-shaped geometry and outputs the stress field. However, as the description of the problem is tailored for aorta models, the network cannot be extended to arbitrary geometry, boundary conditions and loads. We build upon these advances to demonstrate the potential of data-driven techniques for stress field predictions in a moderately more general case.

\section{METHOD}
This work presents a data-driven approach to stress field prediction in 2D cantilevered structures with a linear isotropic elastic material subjected to external loads at the structure's free end. The approach takes as input the structure geometry, external loads, and displacement boundary conditions, and outputs the predicted stress field. Based on the input channel configuration, two deep neural network architectures are proposed: a) SCSNet with a single input channel, and b) StressNet with multiple input channels.

\subsection{Problem Description and Dataset}
\begin{figure}[ht]
	\centering
	\includegraphics[width=0.6\linewidth]{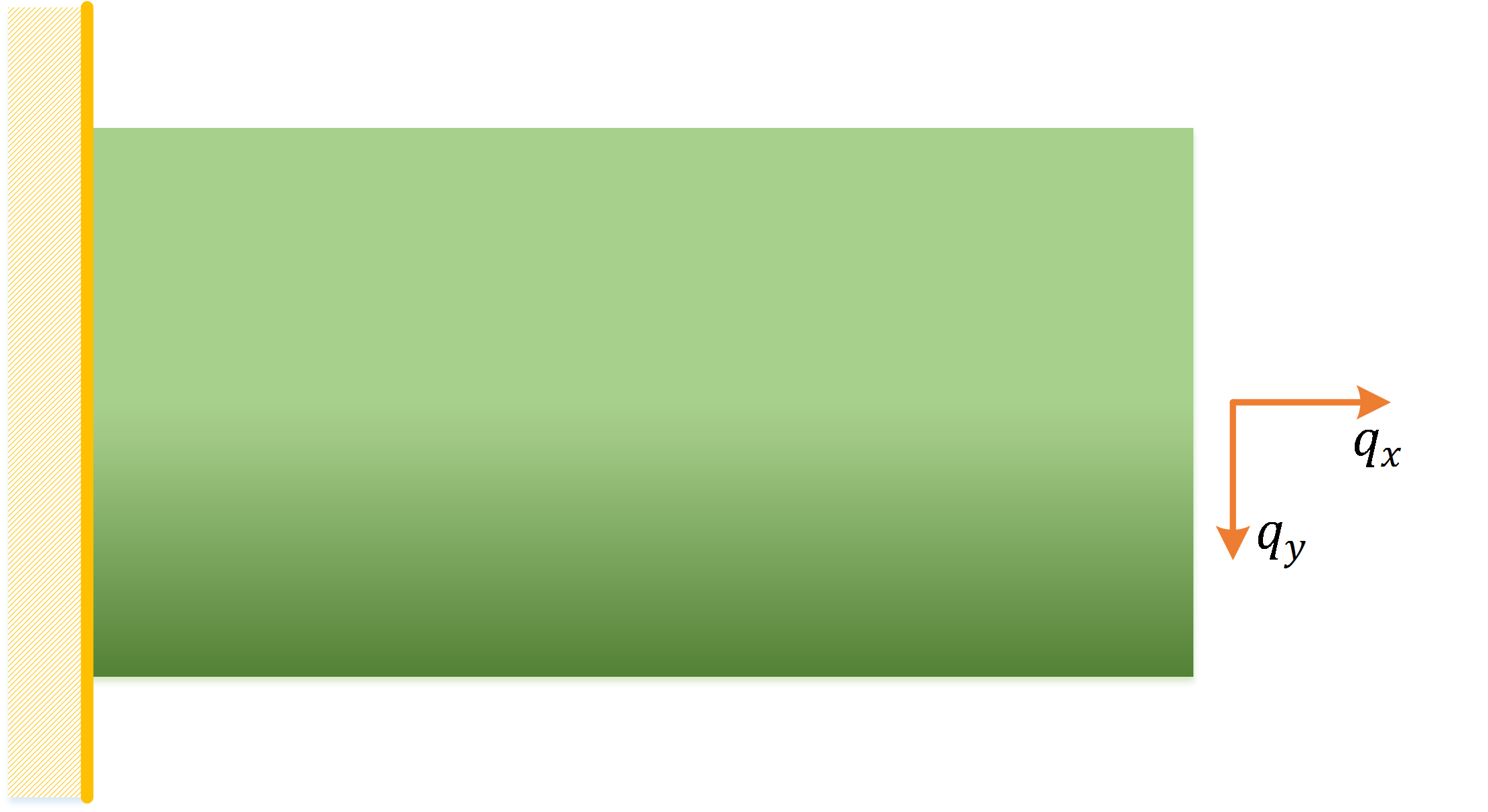}
	\caption{The schematic diagram of a two-dimensional cantilevered structure with linear isotropic material.}
	\label{fig: cantileverBeam}
\end{figure}

Consider the cantilevered structure in Figure \ref{fig: cantileverBeam} composed of a homogeneous and isotropic linear elastic material. The left end of the structure is affixed to the wall, and the right end bears the external static loads. The evenly distributed external loads are applied in both the horizontal ($q_x$) and vertical ($q_y$) directions. For each sample, the loads ($q_x$, $q_y$) are constant due to the static problem, and determined by the resultant force $q$ and its direction $\theta$: $q_x = q\cos{\theta}$ and $q_y = q\sin{\theta}$. The geometry is not limited to a rectangle, but other shapes such as the trapezoid, the trapezoid with curved sides, and all the above structures with holes, are utilized. In total, there are 28 geometries within such six categories with the change of geometry contour, hole shape, size, and location. Theoretically speaking, the eligible cantilevered structures are infinite, so that we cannot enumerate them all. What we select are common structures in mechanical engineering. In addition, the material properties keep unchanged and isotropic for all samples.

This two-dimensional elastic deformation is a plane strain problem. The governing equations consist of the strain-displacement equation (\ref{varepsilon}), compatibility equation (\ref{compatibility}), equilibrium equation (\ref{equilibrium}), and the generalized Hooke's law in Equation (\ref{Hooke}):

\begin{equation}
\varepsilon_{ij}\ =\ \frac{1}{2}\left(\frac{\partial u_i }{\partial x_j } + \frac{\partial u_j }{\partial x_i } \right)\label{varepsilon}
\end{equation}

\begin{equation}
 \frac{1}{2}\left(\frac{\partial^2 \varepsilon_{11} }{\partial x_2^2 } + \frac{\partial^2 \varepsilon_{22}}{\partial x_1^2 } \right) = \frac{\partial^2 \varepsilon_{12} }{\partial x_2 \partial x_1}\label{compatibility}
\end{equation}

\begin{equation}
\frac{\partial \sigma_{ii} }{\partial x_i} + \frac{\partial \sigma_{ij} }{\partial x_j} + f_i = 0\label{equilibrium}
\end{equation}

\begin{equation}
\left[\begin{matrix}a_{11}\\a_{21}\\a_{31}\end{matrix}\right]=\frac{E}{(1-\nu)(1-2\nu)}\left[\begin{matrix}1- \nu&0&0\\0&1-\nu&0\\0&0&1-2 \nu\end{matrix}\right]\left[\begin{matrix}\varepsilon_{11}\\ \varepsilon_{22} \\ \varepsilon_{12}\end{matrix}\right]\label{Hooke}
\end{equation}
where i and j are subscripts with values of 1 or 2, $x_1$ represents the x-axis and $x_2$ represents the y-axis, $u_i$ is the displacement in $x_i$ direction, $\varepsilon_{ij}$ is strain on surface $x_i$ in $x_j$ direction, $\sigma_{ij}$ is stress on surface $x_i$ in $x_j$ direction, $f_i$ is the body force component in $x_i$ direction , $E$ is Young's modulus, and $\nu$ is Poisson's ratio.

\begin{figure}[!htb]
	\centering
	\includegraphics[width=0.61\linewidth]{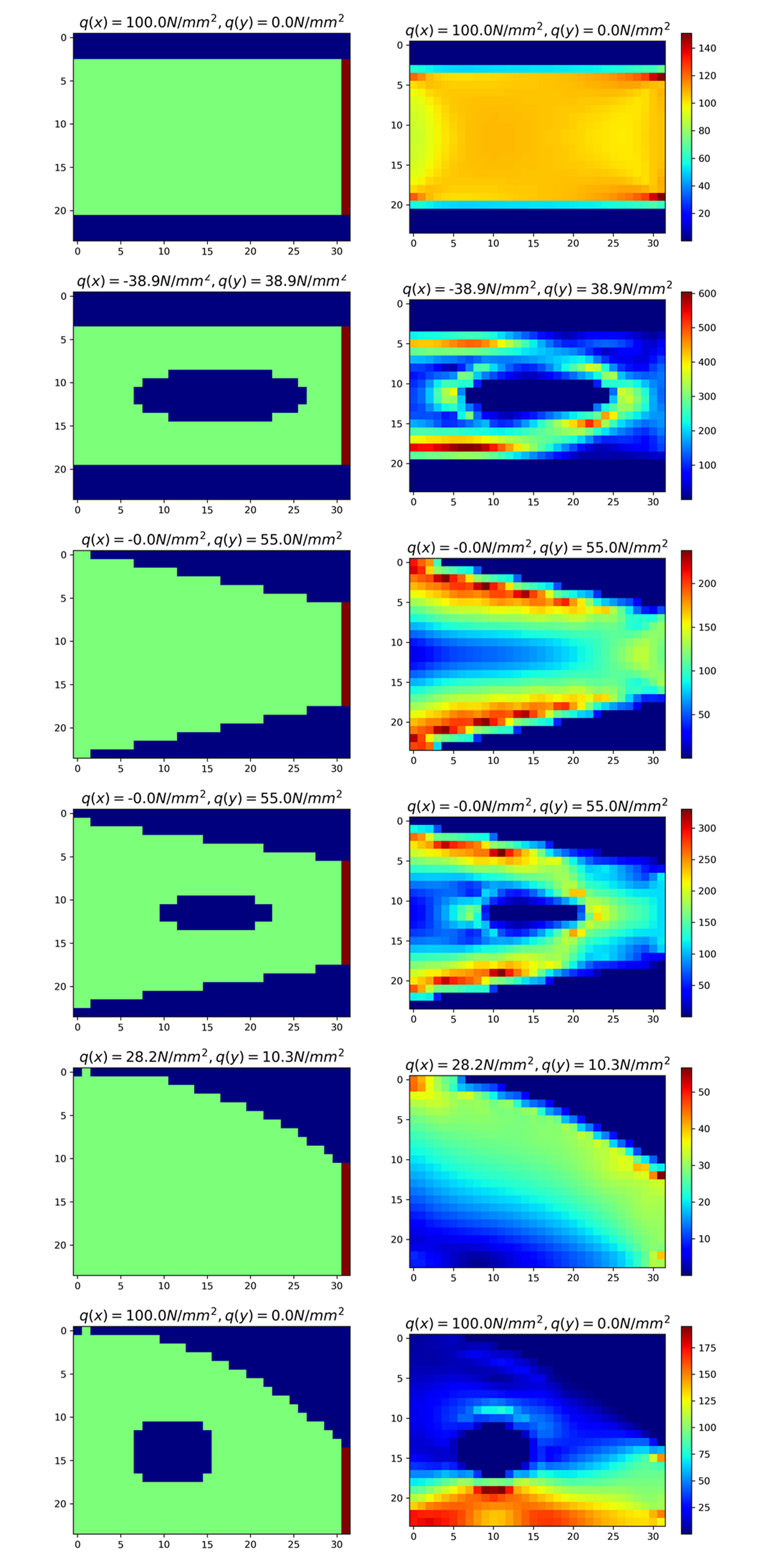}
	\caption{Samples of the dataset computed by FEM. Each row is a FEM sample. Images in the left column are input channels including geometry and load position; Right images are stress fields (Units: mm-MPa-N).
	}
	\label{fig: FEMSample}
\end{figure}

A 2D finite element method (FEM) software SolidsPy \cite{solidspy} is used to generate the training and testing data. The full domain contains $32\times24$ elements. Any element in the domain is a 4-node quadrilateral with a size of $1\times1$ (mm). Randomly selected samples from a total of 120,960 samples are shown in Figure \ref{fig: FEMSample}. Each case is represented as a $32\times24$ image. Images in the left column are the input for the single channel, and the images on the right are the corresponding von Mises ground truth stress fields. For input images, the blue color represents the part of the domain with no material; the green color represents the solid domain; the brown color on the right-hand side of the input data represents the position of the loads. In dataset generation, the resultant load $q$ ranges from 0 to 100 N in an interval of 20 N, and the load direction $\theta$ ranges from 0 to 2$\pi$ in an interval of $\pi/12$. This information delineated with distinct ($q_x$, $q_y$) in the input data. For the ground truth fields, the von Mises stress within the solid domain is plotted as the stress field where the stress is absolute zero within the void domain. Among all computed FEM samples, the von Mises stress varies from 0 to 2,475 MPa with an average of 67.80 MPa.

\begin{figure}[ht]
	\centering
	\includegraphics[width=0.9\linewidth]{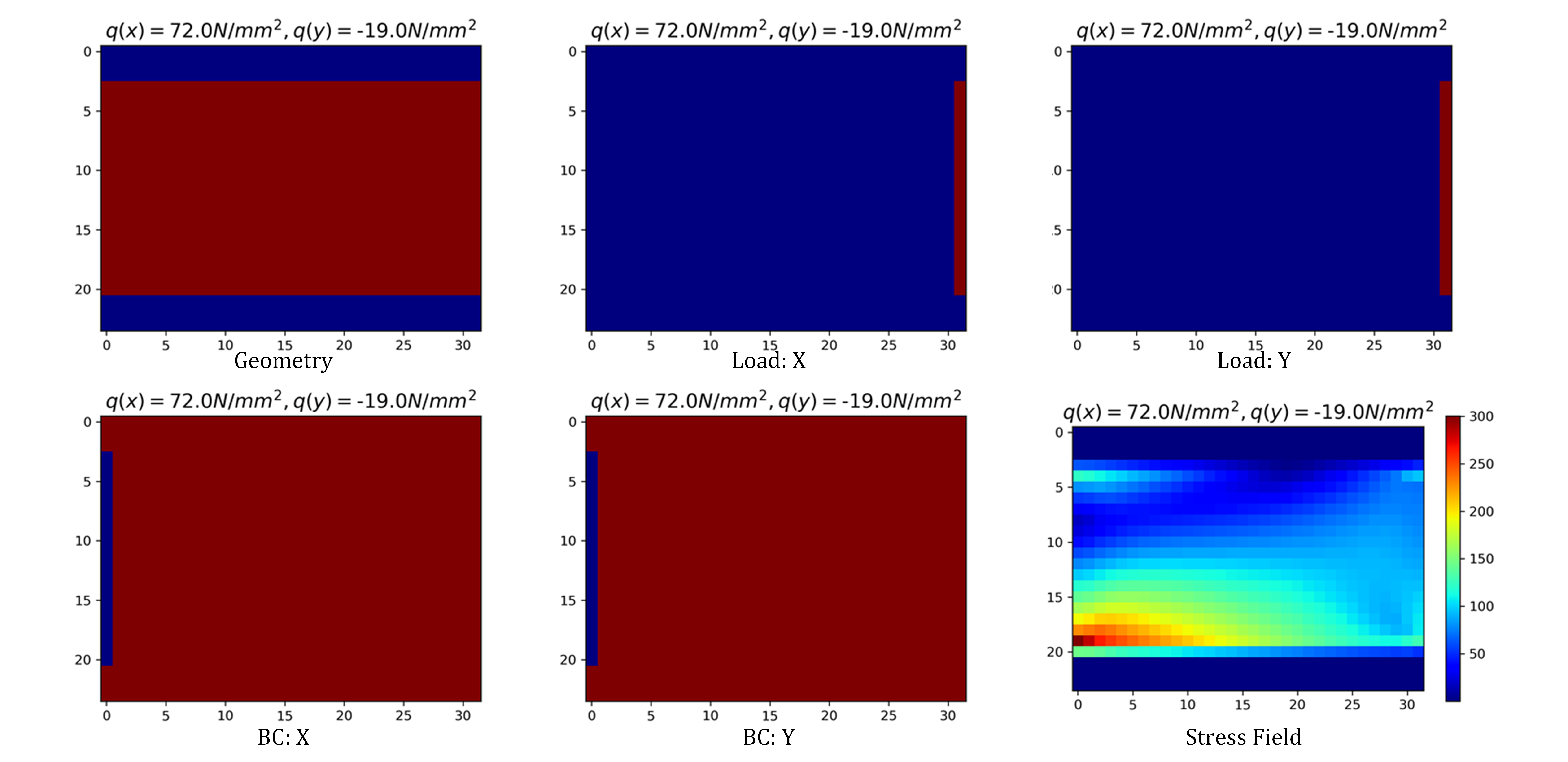}
	\caption{An example of five-channel data representation and the output stress field.}
	\label{fig: MultiChannelSample}
\end{figure}

Based on the computation of FEM, the multi-channel dataset for StressNet can be transformed from the single-channel dataset for SCSNet. As shown in Figure \ref{fig: MultiChannelSample}, all the datasets are encoded as matrices and displayed as colorful images. Firstly, the geometry channel is encoded as a $32\times24$ matrix containing just zeros and ones and shown as a binary-color image. As a single channel geometry, the solid (red denotes 1) and void (blue denotes 0) parts of the domain are distinguished. Secondly, for the two load channels, the component magnitude of the force is located in the matrix at the force applied location (red denotes the force component value, blue denotes 0). Thirdly, for the two displacement boundary condition channels, all displacement-constrained points are delineated in their respective binary matrices (blue denotes -1 and red denotes 0). Finally, the von Mises stress field is also a $32\times24$ matrix and displayed as a colorful image.

\subsection{SCSNet}

\begin{figure}[ht]
	\centering
	\includegraphics[width=0.9\linewidth]{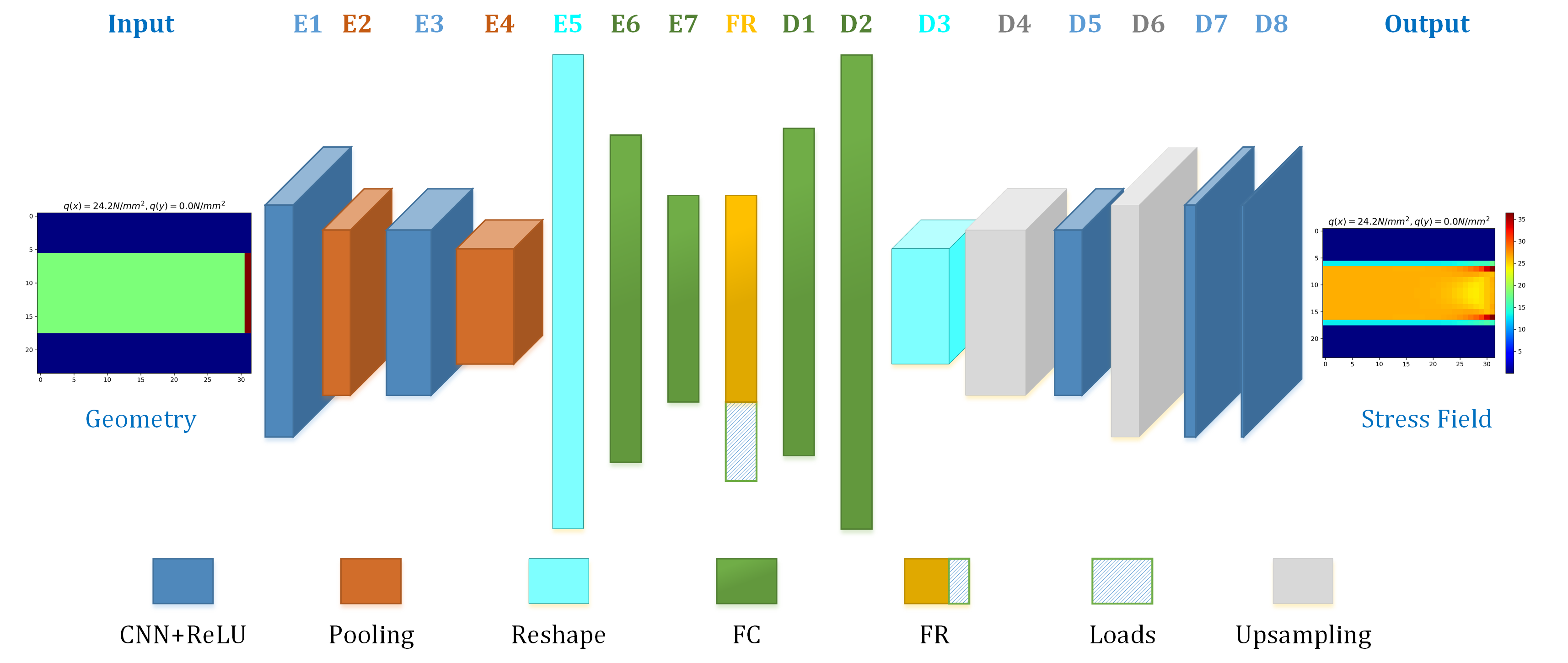}
	\caption{The architecture of SCSNet with a single input channel.}
	\label{fig: BaselineArch}
\end{figure}

As illustrated in Figure \ref{fig: BaselineArch}, a single-channel is proposed to perform the prediction of the von Mises stress field. The input of the model is a matrix including the information of the geometry (blue denotes 0, green denotes 1 and load position (red denotes 2) and shown as a triple-color image. The output of the model is the computed stress field, where each pixel represents the von Mises stress.

SCSNet is a baseline architecture that uses multiple CNN layers in an encoder-decoder structure (convolutional autoencoder). This convolutional autoencoder learns to encode the input in a set of simple signals and then try to reconstruct the input from them \cite{masci2011stacked,geng2015high,holden2015learning}. The encoder network consists of two convolutional layers (E1 and E3) and two max-pooling layers (E2 and E4). Each convolutional layer has a filter with a kernel of $3 \times 3$ and a stride of $1 \times 1$. The padding scheme is zero padding to keep the output image the same size as the input. After a reshape layer E5, and a fully connected (FC) layer E6, we obtain the latent feature representation of geometry at the bottleneck. The load vector ($q_x,q_y$) is concatenated with FR before the new FR is fed into the decoder network. The decoder network is the reverse of the encoder. Upsampling layers take the place of pooling layers for increasing the field resolution. The entire model contains a total of five convolutional layers.

The convolutional layer applies a convolutional operation to the input channels and passes the result to the next layer. CNN can extract distinguishing features from the input images through the scanning and convolutional operation by filter banks \cite{24,25}.

The height, width, and channel of the input image vary through the model: Input image is $24 \times 32 \times 1$; E1 is $24 \times 32 \times 32$; E2 is $12 \times 16 \times 32$; E3 is $13 \times 16 \times 64$; E4 is $6 \times 8 \times 64$; E5 is $3074 \times 1 \times 1$; E6 is $1024 \times 1 \times 1$; E7 is $30 \times 1 \times 1$; FR is $32 \times 1 \times 1$; D1 is $1024 \times 1 \times 1$; D2 is $3074 \times 1 \times 1$; D3 is $6 \times 8 \times 64$; D4 is $12 \times 16 \times 64$; D5 is $12 \times 16 \times 32$; D6 is $24 \times 31 \times 32$; D7 is $24 \times 32 \times 16$; D8-Output image is $24 \times 32 \times 1$.

\subsection{StressNet}
\begin{figure}[ht]
	\centering
	\includegraphics[width=0.9\linewidth]{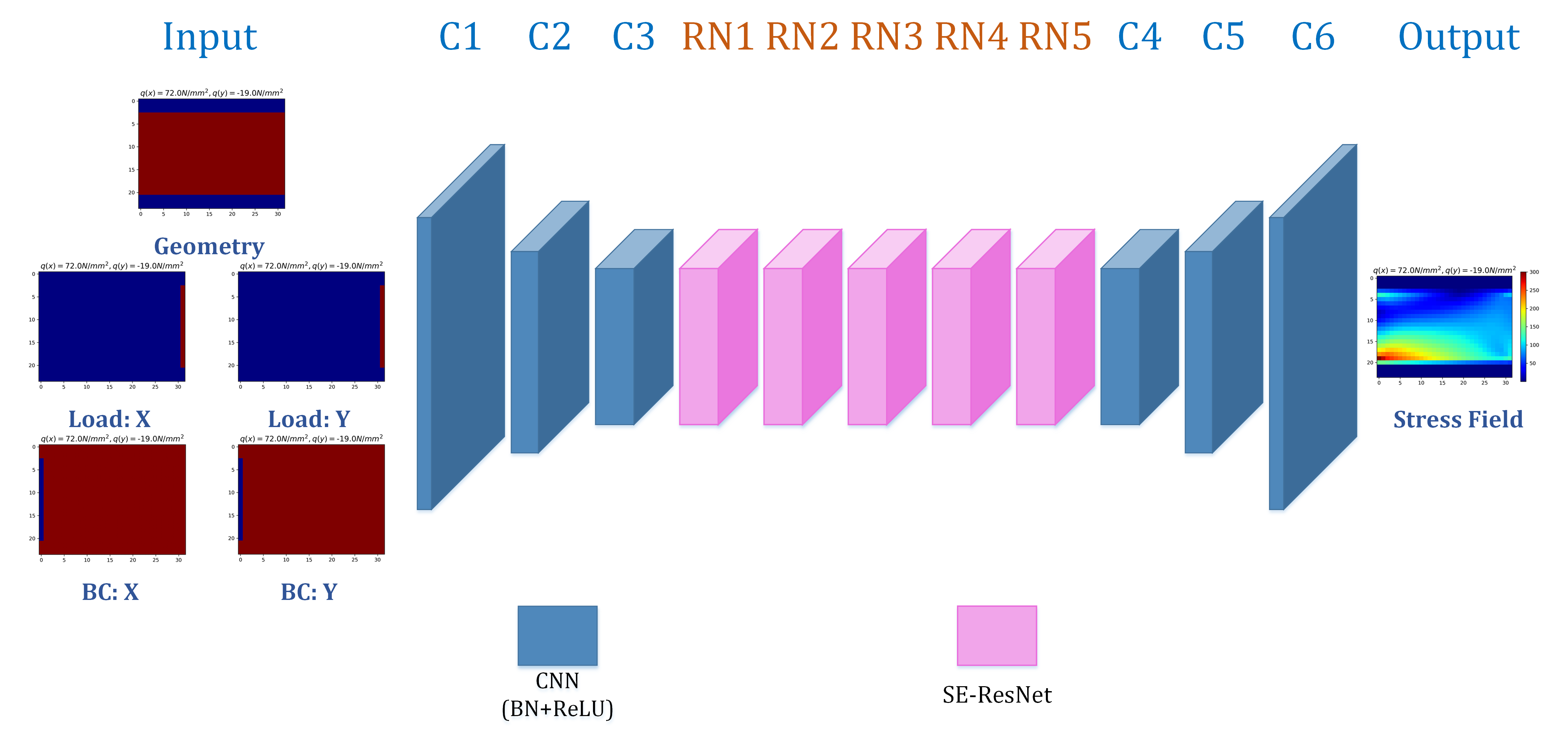}
	\caption{The architecture of StressNet with multiple input channels.}
	\label{fig: StressNet}
\end{figure}

SCSNet has constant CNN layers and can only model simple FEM problems, in which the loads are distributed uniformly and displacement boundary conditions are the same in the X and Y directions. To decrease the prediction error and increase its versatility, we additionally propose a StressNet architecture with multiple input channels as shown in Figure \ref{fig: StressNet}. A downsampling-and-upsampling structure is employed, and five Squeeze-and-Excitation ResNet modules are used between the downsampling and upsampling structures. As mentioned before, ResNet can dynamically select the layer depth by shortcut connections. All convolutional layers are followed by batch normalization and ReLU layers.

The input data contains five channels: 1) geometry, 2) X-component of the load, 3) Y-component of the load, 4) X-component of the displacement boundary condition, 5) Y-component of the displacement boundary condition. For each channel, the information is encoded in a two-dimensional $32 \times 24$ matrix. As with SCSNet, the output data of StressNet is also a single-channel von Mises stress field that is encoded in a $32 \times 24$ matrix and displayed as a colorful image.

Downsampling comprises three convolutional layers (C1, C2, and C3), and upsampling comprises three deconvolutional layers (C4, C5, and C6). Referring to the structure of image transformation networks \cite{26}, we use $9 \times 9$ kernels in the first and last layers (C1 and C6), and $3 \times 3$ kernels in all the other convolutional layers.

\begin{figure}[ht]
	\centering
	\includegraphics[width=0.7\linewidth]{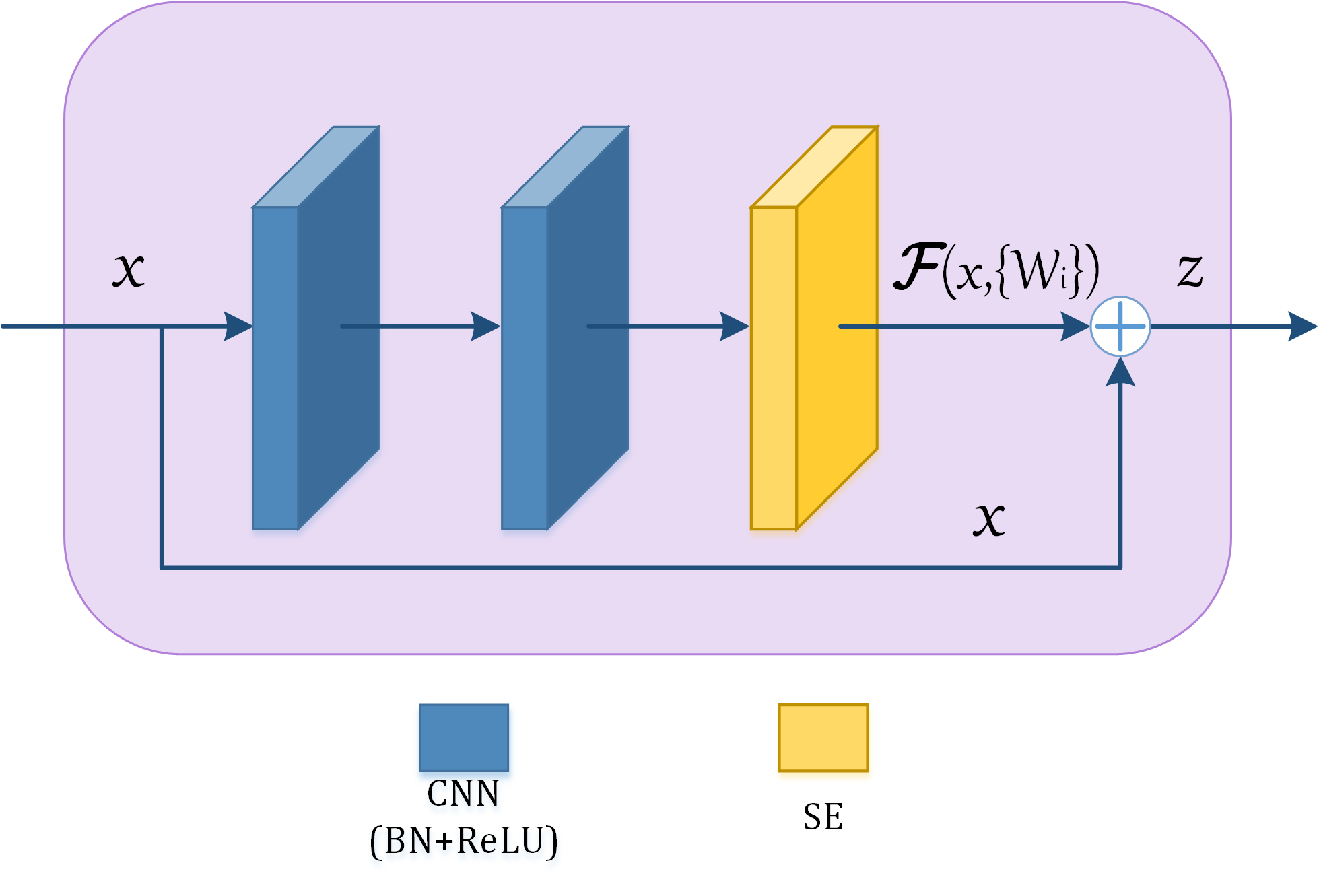}
	\caption{The residual block with a SE block}
	\label{fig: SE-ResNet}
\end{figure}

Residual blocks are used to mimic identical layers to combat the vanishing gradient problem \cite{27}. As shown in Figure \ref{fig: SE-ResNet}, a SE-ResNet module is comprised of two convolutional layers with $3 \times 3$ kernels and one Squeeze-and-Excitation (SE) network block. The output of the SE-ResNet module, as shown in Equation (\eqref{eq_z}), can be conducted by feedforward neural networks with shortcut connections.

\begin{figure}[ht]
	\centering
	\includegraphics[width=0.7\linewidth]{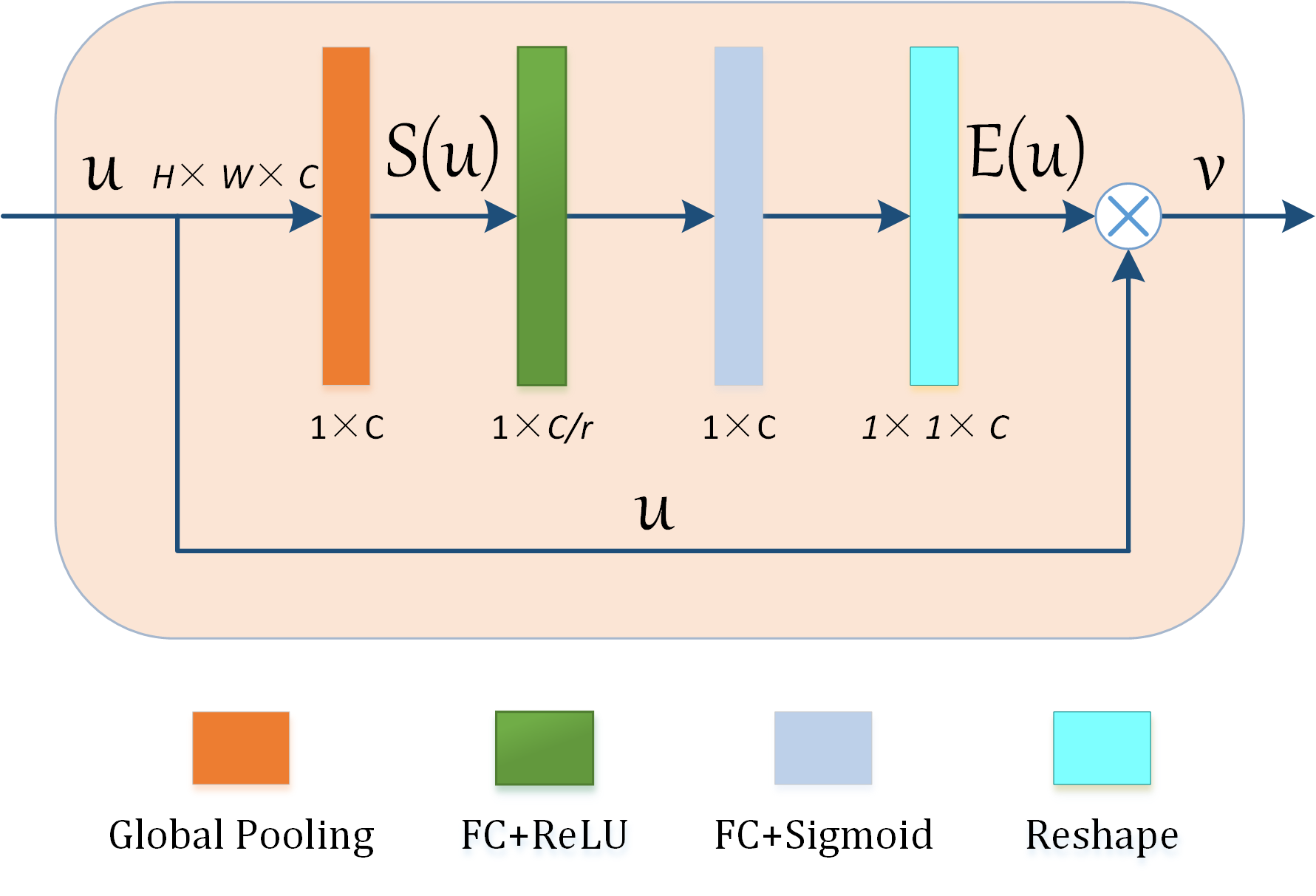}
	\caption{The SE block}
	\label{fig: SE}
\end{figure}

SE blocks, illustrated in Figure \ref{fig: SE} are used inside the SE-ResNet modules to improve the representational capacity of the network by enabling it to perform dynamic channel-wise feature recalibration \cite{28}. The input data $u \in R^{H \times W \times C}$ is shrunk into $\mathbb{S}(u) \in R^C$ through the global average-pooling layer. Then two fully connected layers are employed to downsample (FC+ReLU) and upsample (FC+Sigmoid) the linear array respectively. A reshape operation is conducted to obtain the excitation output data $\mathbb{E}(u)$ that has the same dimension and size as the initial input data $u$. The final output of the block is obtained by a rescaling operation that is the element-wise matrix multiplication, as shown in Equation (\ref{eq_v}).

\begin{equation}
z = \mathbb{F}(x, \{w_i\}) + x\label{eq_z}
\end{equation}

\begin{equation}
v = \mathbb{E}(u) \times u\label{eq_v}
\end{equation}

\subsection{Loss function and metrics}

We use the MSE (mean squared error) and MAE (mean absolute error) to evaluate the model's prediction accuracy. The prediction $\hat{y}$ and ground truth $y$ in our current model are both displayed as $32 \times 24$ resolution images. Before computing MSE and MAE, each predicted stress field is reshaped into 1D-arrays with a length of 768. The reshaped prediction $\hat{y}$ can be expressed as $\hat{y} = (\hat{y_1}, \hat{y_2}, \cdots, \hat{y_n})$, while the reshaped ground truth $y = (y_1, y_2, \cdots, y_n)$. MSE and MAE are respectively represented in Equation (\ref{mse}) and (\ref{mae}).

\begin{equation}
{\rm MSE} = \frac{1}{n} \sum^{n}_{j=1} (y_j-\hat{y_j})^2\label{mse}
\end{equation}
\begin{equation}
{\rm MAE} = \frac{1}{n} \sum^{n}_{j=1} |y_j-\hat{y_j}|\label{mae}
\end{equation}

\noindent where $n=768$ is the total number of elements.

Here we also introduce the mean relative error (MRE) as shown in Equation (\ref{MRE}). MRE is a relative error rate in percentage to measure how close the predictions are to the ground truth.

\begin{equation}
{\rm MRE} = \frac{1}{n} \sum^{n}_{j=1} \frac{|y_j-\hat{y_j}|}{\epsilon + {\rm max}(y_j,\hat{y_j})} \times 100 \% \label{MRE}
\end{equation}
where $\epsilon$ is a smoothing term that avoids division by zero (this article takes $\epsilon = 0.01$).

Based on the Cauchy-Schwartz inequality, ${\rm MSE} \leq {\rm MAE^2}$. MSE has a tendency to be increasingly larger than ${\rm MAE}^2$ with an increase of testing sample, and is more sensitive to data variance. In our approach, MSE is used for the training loss, while all three error measures are used to quantify the prediction performance.

\section{RESULTS AND DISCUSSIONS}

All code is written in TensorFlow and run on an NVIDIA GeForce GTX 1080Ti GPU. Adam optimization algorithm with an exponentially decaying learning rate is used for training, and the batch size is set to 256, which is the allowed maximum size according to the GPU memory. The training and testing results show that both the two architectures are stable and converged reliably. Under our experiment scenarios, it takes 1.56 seconds for SCSNet to render all the 120,960 FEM samples, and 10.4 seconds for StressNet. The FEM software takes approximately ten hours (on Intel i7-6500U CPU) to accomplish the FEM computation on the same number of 2D problems. 

\subsection{Accuracy and performance}
\begin{figure}[ht]
	\centering
	\includegraphics[width=0.8\linewidth]{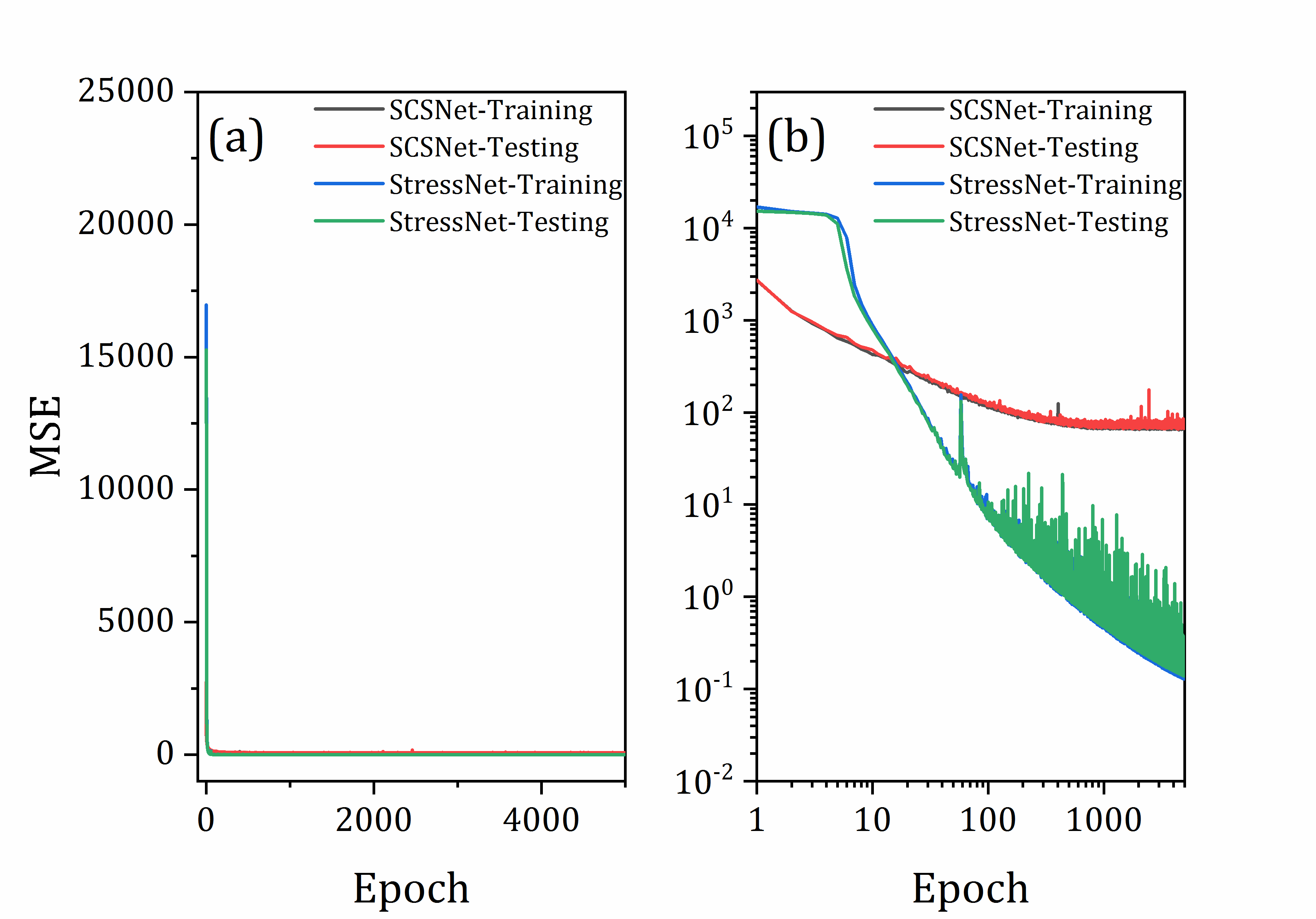}
	\caption{MSE curves on training and testing data of two architectures. (a) is shown in arithmetic coordinates; (b) is shown in logarithmic coordinates.}
	\label{fig: MSETwoModel}
\end{figure}

\begin{figure}[ht]
	\centering
	\includegraphics[width=0.8\linewidth]{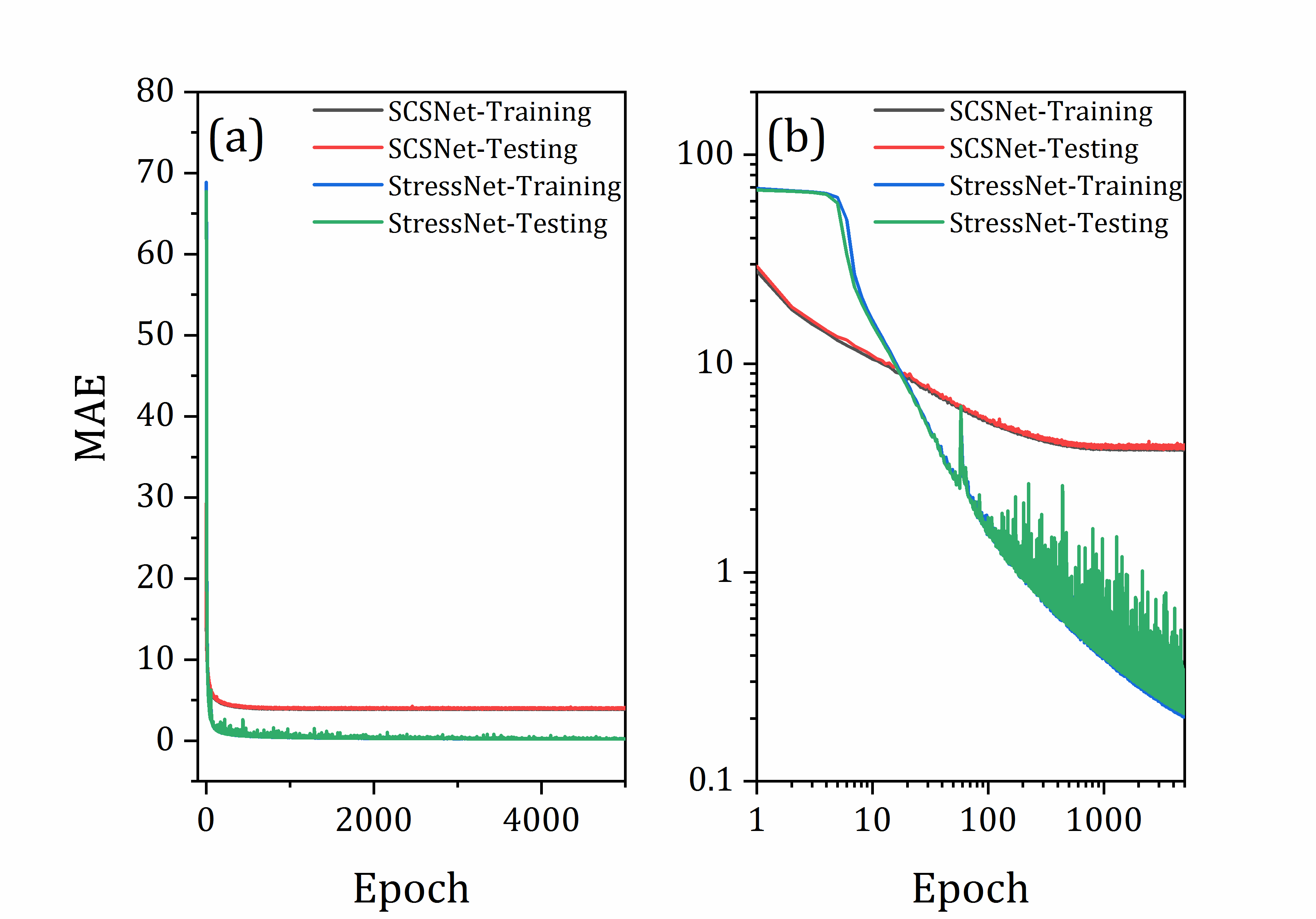}
	\caption{MAE curves on training and testing data of two architectures. (a) is shown in arithmetic coordinates; (b) is shown in logarithmic coordinates.}
	\label{fig: MAETwoModel}
\end{figure}

In this experiment, we train and evaluate our models using the whole dataset. The training data size is 100,000, and the separate testing data size is 20,960. Figure \ref{fig: MSETwoModel} shows MSE loss as a function of epochs. Chart (a) is in arithmetic coordinates, and chart (b) is in logarithmic coordinates. Chart (a) shows that all the four MSE curves decline rapidly in the first fifty epochs, and then begin to flatten. From chart (b), it can be seen that SCSNet preserves a nearly constant order of magnitude after 1000 epochs. By contrast, StressNet continues to decrease with more training, even after 5000 epochs. StressNet has a significantly smaller MSE than SCSNet in the end. Figure \ref{fig: MAETwoModel} shows the MAE loss on training data and testing data of the two architectures. The trend in the MAE loss is nearly identical to that of MSE. StressNet has much better accuracy than SCSNet. The other performance metrics are summarized in Table \ref{tab: metrics}. It can be seen that MRE of StressNet is just 2.04\% for testing, which we deem acceptable for stress fields prediction.

\begin{figure*}[ht]
	\centering
 \includegraphics[trim = 0in 0in 0in 0in, clip, width=\textwidth]{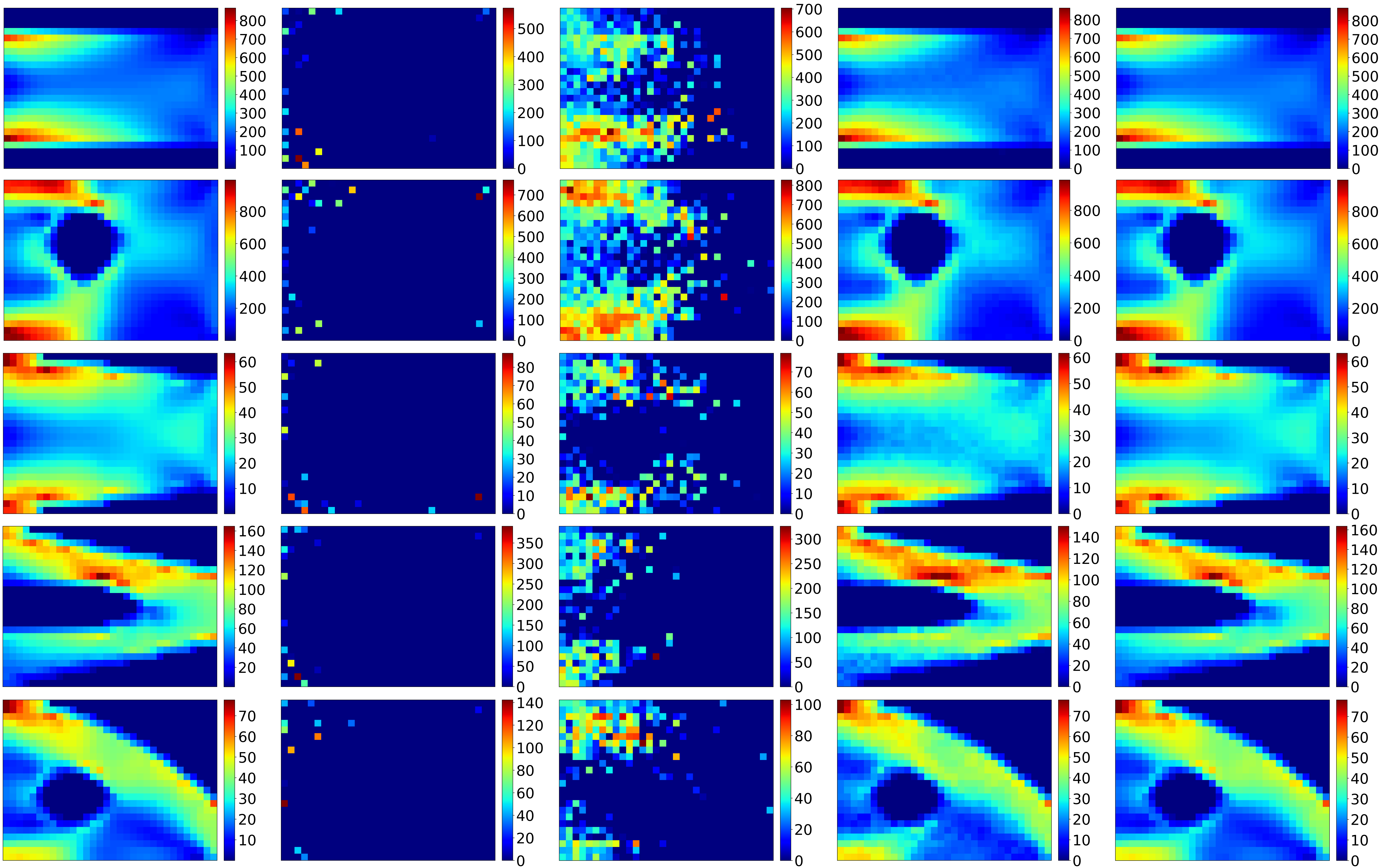}
	\caption{Computed stress fields by StressNet: ground truth and evolutive predictions. Each row denotes a randomly selected sample form the whole datasets. From left to right: 1) ground truth; 2) epoch = 1; 3) epoch = 10; 4) epoch = 100; 5) epoch = 5000. The ratio of training data and testing data is 100,000/20,960.
	}
	\label{fig: evolutive predictions}
\end{figure*}

The evolutive predictions by StressNet with the increasing epoch during the training are plotted in Figure \ref{fig: evolutive predictions}. Each row denotes a randomly selected sample from the whole datasets. The first column is the ground truth. Images starting from the second column, in order, are evolutive predictions of the von Mises stress fields. Predictions in the first epoch just contain a few non-zero scattering points, and then in the tenth epoch have a rudimentary form, especially on the left half. Since from the hundredth epoch, the predictions have no visual distinction with the ground truth.

\begin{table}[!htb]
\centering
\caption{\label{tab: metrics} \textbf{Error metrics.} Epoch = 5000, $\bar{y} = 67.80$. SCS denotes SCSNet, and SN denotes StressNet.}

\begin{tabular}{|c|c|c|c|c|c|c|}
\hline
Metric & \multicolumn{2}{c|}{MSE (${\rm MPa}^2$)} & \multicolumn{2}{c|}{MAE (MPa)} & \multicolumn{2}{c|}{MRE (\%)} \\ 
\hline
Model & SCS & SN & SCS & SN & SCS & SN \\ 
\hline
Training & 83.63 & 0.14& 4.28 & 0.22 & 10.40 & 1.99 \\ 
\hline
Testing & 84.07 & 0.15 & 4.30 & 0.23 & 10.43 & 2.04 \\ 
\hline
\end{tabular}
\end{table}

\subsection{Effect of the training data size on performance}
\begin{figure}[!ht]
	\centering
	\includegraphics[width=0.8\linewidth]{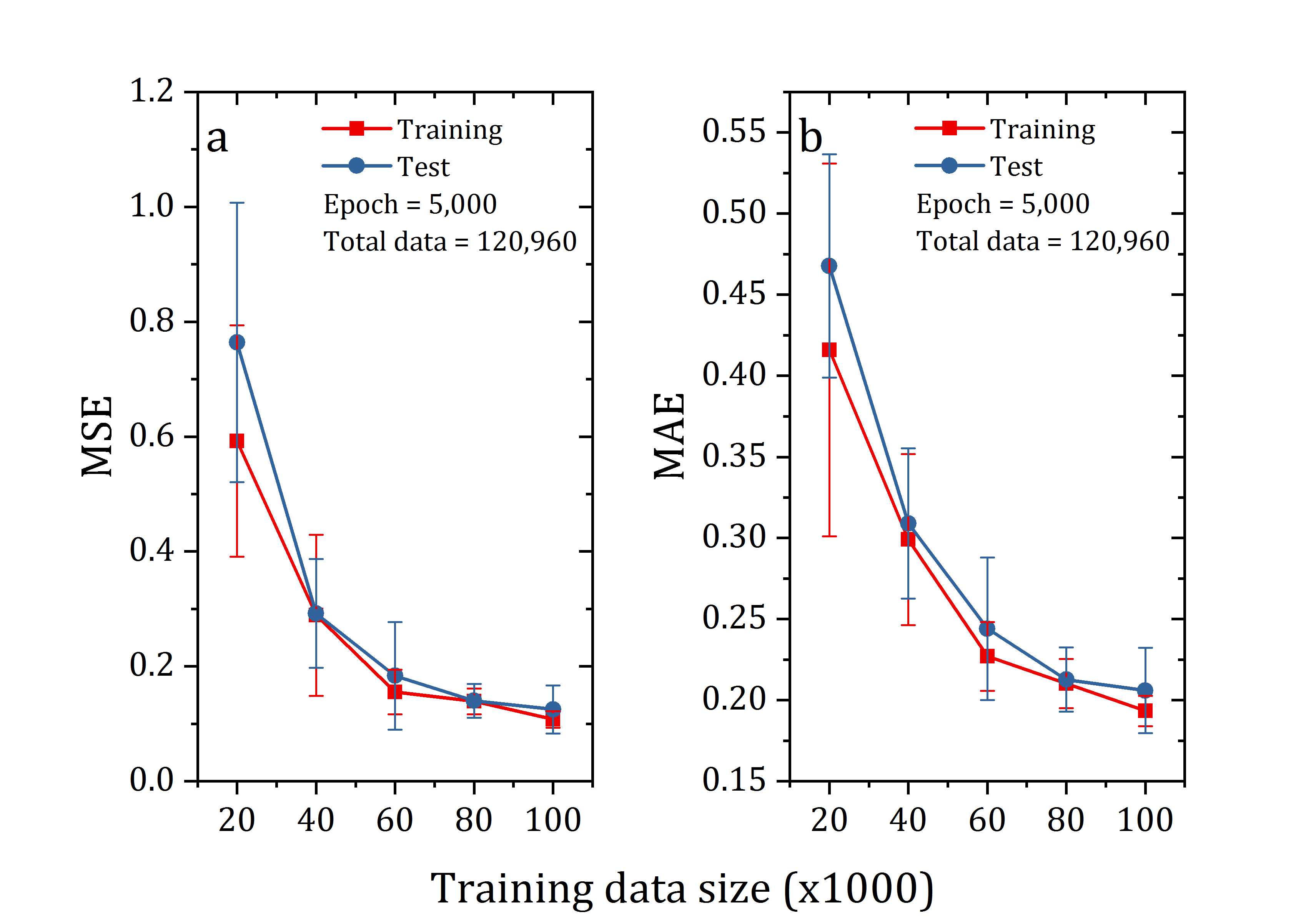}
	\caption{Effect of training data size on the prediction performance of StressNet.}
	\label{fig: MAEMSEdata}
\end{figure}

\begin{figure*}[!tp]
	\centering
 \includegraphics[width=0.9\linewidth]{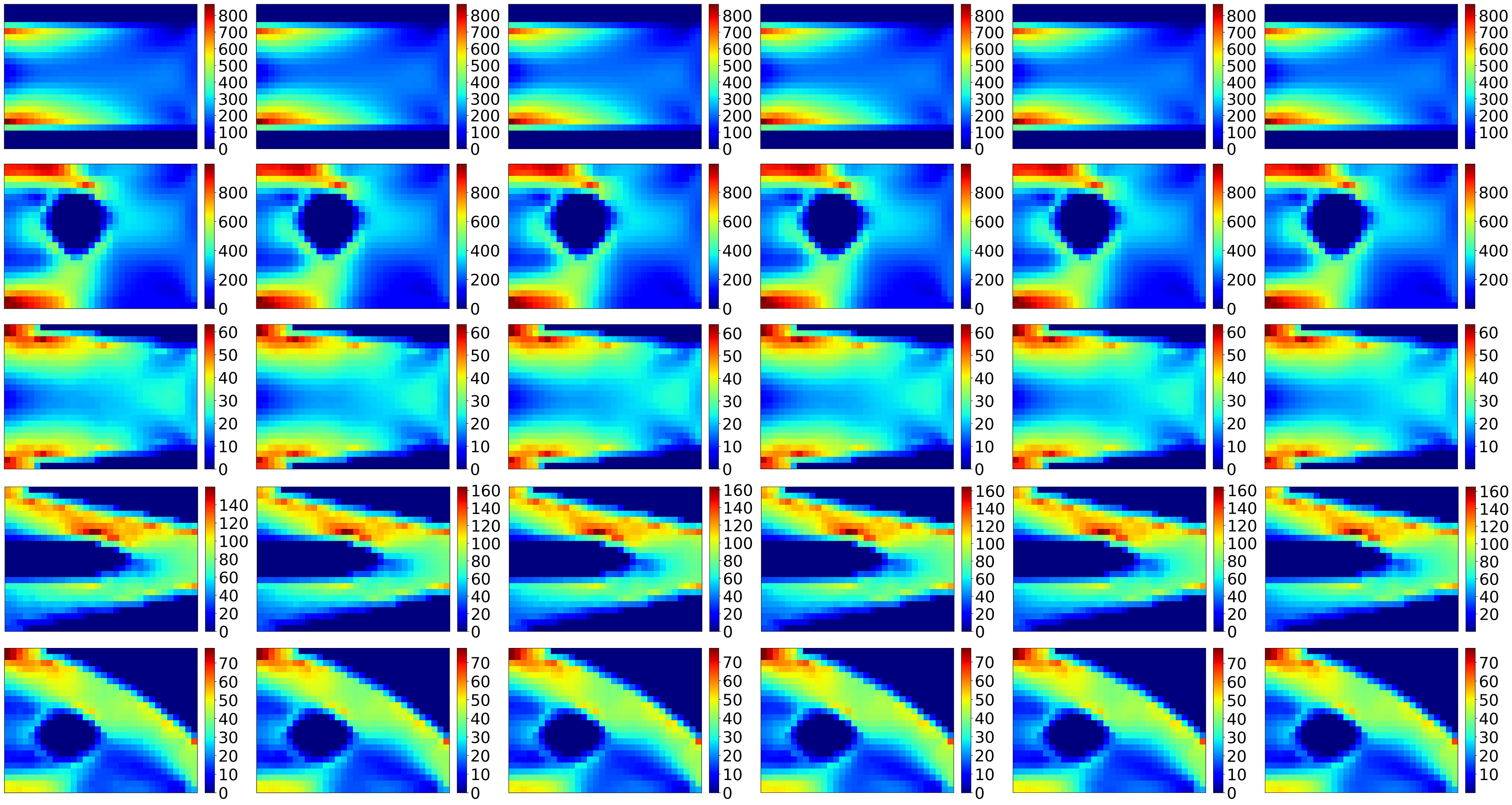}
	\caption{Computed stress fields with different training data sizes by StressNet. From left to right, the training data size is: 1) 20,000; 2) 40,000; 3) 60,000; 4) 80,000; 5) 100,000; 6) ground truths.}
	\label{fig: data}
\end{figure*}

Besides using all the data for training and testing with the ratio of 100,000/20,960, we also reduce the training data size from 100,000 to 20,000 to demonstrate the effect of training size. Figure \ref{fig: MAEMSEdata} shows the variations of MSE and MAE for both training and testing with different training sizes. The variations of MSE and MAE follow similar trends. With the increase in the training data, both MSE and MAE decline gradually. The standard deviation across multiple runs at a fixed training data size (error bar) also decreases with increasing training data. MSE and MAE of the testing data are expectantly slightly larger than those of the training data. It means that the larger training data, the better prediction performance. Similarly, we randomly plot the computed stress fields with different training data size as shown in Figure \ref{fig: MAEMSEdata}. By using different sizes of the training data, StressNet training is terminated at the five thousandth epoch. For each sample in Figure \ref{fig: data}, the stress fields predicted by StressNet look like the same with different training data sizes. However, they have different color bars, which means the outputs are similar but still different.

\subsection{Prediction of maximum stress}
As a measure of how well the results of StressNet can be used for predicting failure, we also examine how well StressNet can predict the maximum von Mises stress obtained in the ground truth. An analysis of the testing data shows a coefficient of determination $R^{2}$ of 0.99. This implies that StressNet is able to predict the maximum stress with significant accuracy.

\subsection{Effect of the hierarchical nature of deep learning on prediction accuracy}
\begin{figure}[!htb]
	\centering
	\includegraphics[width=0.9\linewidth]{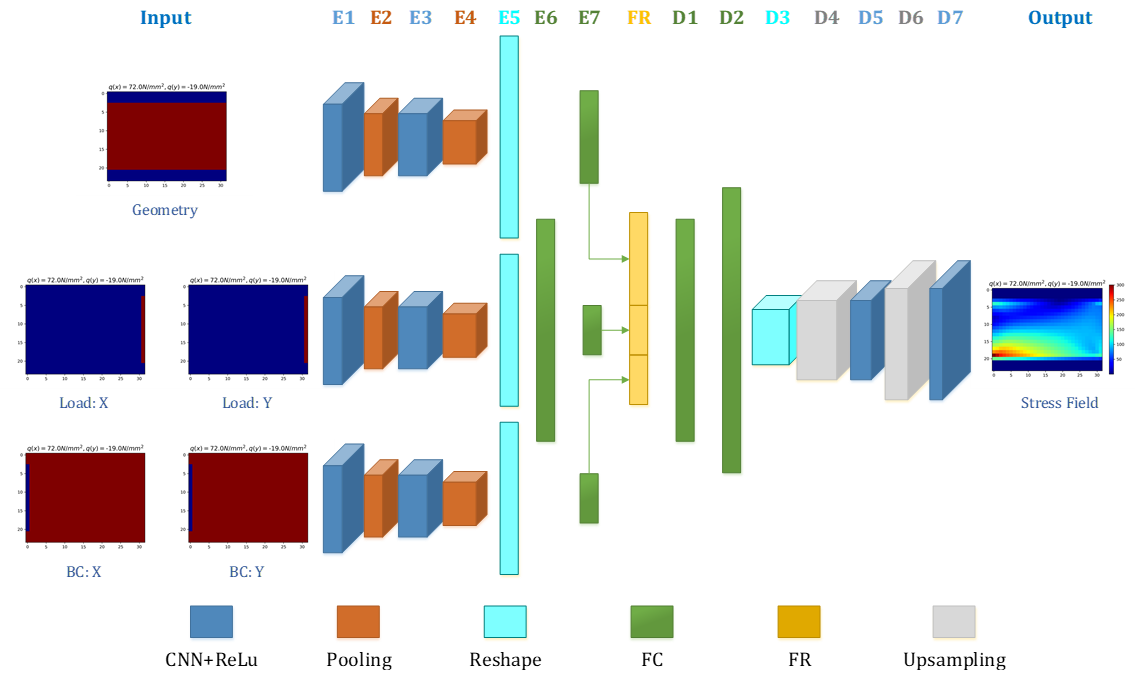}
	\caption{An inbetween architecture with representation fusion and multiple input channels.}
	\label{fig: intermediate}
\end{figure}

In addition to the two architectures described above, we also deployed and studied an inbetween architecture shown in Figure \ref{fig: intermediate} which keeps the SCSNet structure but uses StressNet's multiple channeled inputs. The five channels are classified into three categories: geometry, loads and boundary conditions similar to the way described for StressNet. All three encoders are independent and parallel to each other. After the fully connected layers, three feature representations are combined in turn. Then the combined FR is decoded through reverse CNN layers. Such an intermediate model consists of eight convolutional layers totally, which is far less than StressNet.

We trained this model on the same GPU, and the training result shows that this model has no improvement in training accuracy relative to SCSNet. This supports our hypothesis that as the hierarchical architecture becomes deeper as it does with StressNet, its prediction becomes more accurate, although this deepening has to be judicious in light of the image resolution, training size, and data variability. Additionally, the input channel configuration may not have a significant impact on prediction accuracy.

Comparatively speaking, SCSNet can reduce the training time, and reach an acceptable accuracy where the mean relative error is 10.40\%. In addition, due to its fully convolutional network architecture, it is able to account for large variations in the size of the 2D FEM data. This makes StressNet a great alternative to classical FEM once the training data is sufficient to train the network.

\section{CONCLUSION}
In this paper, we present an end-to-end deep learning based approach for stress field prediction in cantilevered structures. Two architectures are implemented: a single-channel stress prediction neural network (SCSNet), and a multi-channel stress prediction neural network (StressNet) with SE-ResNet modules. A 2D FEM software SolidsPy is used to generate the training and testing data, which contains a total of 120,960 FEM samples.

Both architectures are stable and converged reliably in training and testing. MSE and MAE results show that StressNet can obtain higher accuracy than SCSNet. The mean relative error of the StressNet model is just 2.04\% for testing with respect to the ground truth. The effect of training data size on performance is also studied. With the increase in the training data size, both MSE and MAE decline gradually. However, as the magnitude of MAE is relatively small with respect to the magnitudes of the stress fields, all the computed stress fields across different training data size are desirably very similar.

The effect of the hierarchical nature of the deep network on prediction accuracy is studied. The results indicate that the input channel number does not have a significant effect on the prediction accuracy. As hierarchical architecture becomes deeper, its prediction becomes more accurate. For better prediction accuracy, StressNet is a reasonable choice. It encodes and preserves more information of the input and enables a richer set of non-linear operations due to its architecture.

Future work will be to improve the network to an increasingly more general method using generative deep learning to alleviate the need for extensive coverage of the input geometry, boundary conditions, and loads.

\bibliographystyle{unsrt}


\bibliography{references}

\begin{thebibliography}{10}

\bibitem{2}
Abhijit~Guha Roy, Sailesh Conjeti, Sri Phani~Krishna Karri, Debdoot Sheet, Amin
  Katouzian, Christian Wachinger, and Nassir Navab.
\newblock Relaynet: retinal layer and fluid segmentation of macular optical
  coherence tomography using fully convolutional networks.
\newblock {\em Biomed. Opt. Express}, 8(8):3627--3642, Aug 2017.

\bibitem{3}
Arvind~T. {Mohan} and Datta~V. {Gaitonde}.
\newblock {A Deep Learning based Approach to Reduced Order Modeling for
  Turbulent Flow Control using LSTM Neural Networks}.
\newblock {\em arXiv e-prints}, page arXiv:1804.09269, Apr 2018.

\bibitem{4}
Amir~Barati Farimani, Joseph Gomes, and Vijay~S. Pande.
\newblock Deep learning the physics of transport phenomena.
\newblock {\em CoRR}, abs/1709.02432, 2017.

\bibitem{5}
Byungsoo Kim, Vinicius~C. Azevedo, Nils Thuerey, Theodore Kim, Markus~H. Gross,
  and Barbara Solenthaler.
\newblock Deep fluids: {A} generative network for parameterized fluid
  simulations.
\newblock {\em CoRR}, abs/1806.02071, 2018.

\bibitem{6}
Nobuyuki Umetani.
\newblock Exploring generative 3d shapes using autoencoder networks.
\newblock pages 1--4, 11 2017.

\bibitem{7}
Yonggyun Yu, Taeil Hur, and Jaeho Jung.
\newblock Deep learning for topology optimization design.
\newblock {\em CoRR}, abs/1801.05463, 2018.

\bibitem{8}
Wentai Zhang, Haoliang Jiang, Zhangsihao Yang, Soji Yamakawa, Kenji Shimada,
  and Levent~Burak Kara.
\newblock Data-driven upsampling of point clouds.
\newblock {\em CoRR}, abs/1807.02740, 2018.

\bibitem{9}
Erva Ulu, Rusheng Zhang, Mehmet~Ersin Yumer, and Levent~Burak Kara.
\newblock A data-driven investigation and estimation of optimal topologies
  under variable loading configurations.
\newblock In {\em Computational Modeling of Objects Presented in Images.
  Fundamentals, Methods, and Applications}, pages 387--399. Springer
  International Publishing, 2014.

\bibitem{14}
Garrett~B. Goh, Nathan~O. Hodas, and Abhinav Vishnu.
\newblock Deep learning for computational chemistry.
\newblock {\em Journal of Computational Chemistry}, 38(16):1291--1307, 2017.

\bibitem{15}
Andreas {Mardt}, Luca {Pasquali}, Hao {Wu}, and Frank {No{\'e}}.
\newblock {VAMPnets for deep learning of molecular kinetics}.
\newblock {\em Nature Communications}, 9:5, Jan 2018.

\bibitem{16}
Gr{\'e}goire {Montavon}, Matthias {Rupp}, Vivekanand {Gobre}, Alvaro
  {Vazquez-Mayagoitia}, Katja {Hansen}, Alexand~re {Tkatchenko}, Klaus-Robert
  {M{\"u}ller}, and O.~{Anatole von Lilienfeld}.
\newblock {Machine learning of molecular electronic properties in chemical
  compound space}.
\newblock {\em New Journal of Physics}, 15:095003, Sep 2013.

\bibitem{17}
Gareth~A. Tribello, Michele Ceriotti, and Michele Parrinello.
\newblock A self-learning algorithm for biased molecular dynamics.
\newblock {\em Proceedings of the National Academy of Sciences},
  107(41):17509--17514, 2010.

\bibitem{10}
Maziar Raissi, Paris Perdikaris, and George Karniadakis.
\newblock Multistep neural networks for data-driven discovery of nonlinear
  dynamical systems.
\newblock 01 2018.

\bibitem{18}
Peter Broecker, Juan Carrasquilla, Roger~G. Melko, and Simon Trebst.
\newblock Machine learning quantum phases of matter beyond the fermion sign
  problem.
\newblock In {\em Scientific Reports}, 2017.

\bibitem{19}
Kristof~T. {Sch{\"u}tt}, Farhad {Arbabzadah}, Stefan {Chmiela}, Klaus~R.
  {M{\"u}ller}, and Alexandre {Tkatchenko}.
\newblock {Quantum-chemical insights from deep tensor neural networks}.
\newblock {\em Nature Communications}, 8:13890, Jan 2017.

\bibitem{21}
Amir {Barati Farimani}, Joseph {Gomes}, Rishi {Sharma}, Franklin~L. {Lee}, and
  Vijay~S. {Pande}.
\newblock {Deep Learning Phase Segregation}.
\newblock {\em arXiv e-prints}, page arXiv:1803.08993, Mar 2018.

\bibitem{point_cloud}
Wentai Zhang, Haoliang Jiang, Zhangsihao Yang, Soji Yamakawa, Kenji Shimada,
  and Levent~Burak Kara.
\newblock Data-driven upsampling of point clouds.
\newblock {\em CoRR}, abs/1807.02740, 2018.

\bibitem{FEA_modelUpdate1}
R.I. Levin and N.A.J. Lieven.
\newblock Dynamic finite element model updating using neural networks.
\newblock {\em Journal of Sound and Vibration}, 210(5):593 -- 607, 1998.

\bibitem{FEA_modelUpdate2}
M.J. Atalla and D.J. Inman.
\newblock On model updating using neural networks.
\newblock {\em Mechanical Systems and Signal Processing}, 12(1):135 -- 161,
  1998.

\bibitem{NN_FAE}
A.A. Javadi and T~P.~Tan.
\newblock Neural network for constitutive modelling in flnite element analysis.
\newblock {\em Computer Assisted Mechanics and Engineering Sciences}, 10, 01
  2003.

\bibitem{23}
Atsuya Oishi and Genki Yagawa.
\newblock Computational mechanics enhanced by deep learning.
\newblock {\em Computer Methods in Applied Mechanics and Engineering}, 327:327
  -- 351, 2017.
\newblock Advances in Computational Mechanics and Scientific Computation—the
  Cutting Edge.

\bibitem{NN_plausibilityCheck}
Tobias Spruegel, Tina Schröppel, and Sandro Wartzack.
\newblock Generic approach to plausibility checks for structural mechanics with
  deep learning.
\newblock 08 2017.

\bibitem{surrogate_FEA}
Liang Liang, Minliang Liu, Caitlin Martin, and Wei Sun.
\newblock A deep learning approach to estimate stress distribution: a fast and
  accurate surrogate of finite-element analysis.
\newblock {\em Journal of The Royal Society Interface}, 15, 01 2018.

\bibitem{1}
Yann LeCun, Y~Bengio, and Geoffrey Hinton.
\newblock Deep learning.
\newblock {\em Nature}, 521:436--44, 05 2015.

\bibitem{simonyan2014very}
Karen Simonyan and Andrew Zisserman.
\newblock Very deep convolutional networks for large-scale image recognition.
\newblock {\em arXiv preprint arXiv:1409.1556}, 2014.

\bibitem{27}
Kaiming He, Xiangyu Zhang, Shaoqing Ren, and Jian Sun.
\newblock Deep residual learning for image recognition.
\newblock {\em CoRR}, abs/1512.03385, 2015.

\bibitem{jenkins1995neural}
WM~Jenkins.
\newblock Neural network-based approximations for structural analysis.
\newblock In {\em Developments in Neural Networks and Evolutionary Computing
  for Civil and Structural Engineering}, pages 25--35. Civil-Comp Press
  Edinburgh, 1995.

\bibitem{waszczyszyn2001neural}
Zenon Waszczyszyn and Leonard Ziemia{\'n}ski.
\newblock Neural networks in mechanics of structures and materials--new results
  and prospects of applications.
\newblock {\em Computers \& Structures}, 79(22-25):2261--2276, 2001.

\bibitem{goh1995multivariate}
ATC Goh, KS~Wong, and BB~Broms.
\newblock Multivariate modelling of fem data using neural networks.
\newblock {\em CIVIL-COMP95 Developments in Neural Networks and Evolutionary
  Computing for Civil and Structural Engineering}, pages 59--64, 1995.

\bibitem{abendroth2003determination}
Martin Abendroth and Meinhard Kuna.
\newblock Determination of deformation and failure properties of ductile
  materials by means of the small punch test and neural networks.
\newblock {\em computational materials Science}, 28(3-4):633--644, 2003.

\bibitem{wu1992use}
X~Wu, J~Ghaboussi, and JH~Garrett~Jr.
\newblock Use of neural networks in detection of structural damage.
\newblock {\em Computers \& structures}, 42(4):649--659, 1992.

\bibitem{zang2001structural}
C~Zang and M~Imregun.
\newblock Structural damage detection using artificial neural networks and
  measured frf data reduced via principal component projection.
\newblock {\em Journal of sound and vibration}, 242(5):813--827, 2001.

\bibitem{tsou1994structural}
Poyu Tsou and M-HH Shen.
\newblock Structural damage detection and identification using neural networks.
\newblock {\em AIAA journal}, 32(1):176--183, 1994.

\bibitem{huber1999determination}
N~Huber and Ch~Tsakmakis.
\newblock Determination of constitutive properties from spherical indentation
  data using neural networks. part ii: plasticity with nonlinear isotropic and
  kinematic hardening.
\newblock {\em Journal of the Mechanics and Physics of Solids},
  47(7):1589--1607, 1999.

\bibitem{zhang2003artificial}
Z~Zhang and K~Friedrich.
\newblock Artificial neural networks applied to polymer composites: a review.
\newblock {\em Composites Science and technology}, 63(14):2029--2044, 2003.

\bibitem{settgast2019constitutive}
Christoph Settgast, Martin Abendroth, and Meinhard Kuna.
\newblock Constitutive modeling of plastic deformation behavior of open-cell
  foam structures using neural networks.
\newblock {\em Mechanics of Materials}, 2019.

\bibitem{liu2019estimation}
Minliang Liu, Liang Liang, and Wei Sun.
\newblock Estimation of in vivo constitutive parameters of the aortic wall
  using a machine learning approach.
\newblock {\em Computer Methods in Applied Mechanics and Engineering},
  347:201--217, 2019.

\bibitem{yildiz2003integrated}
AR~Yildiz, Nursel {\"O}zt{\"u}rk, Necmettin Kaya, and Ferruh {\"O}zt{\"u}rk.
\newblock Integrated optimal topology design and shape optimization using
  neural networks.
\newblock {\em Structural and Multidisciplinary Optimization}, 25(4):251--260,
  2003.

\bibitem{papadrakakis2002reliability}
Manolis Papadrakakis and Nikos~D Lagaros.
\newblock Reliability-based structural optimization using neural networks and
  monte carlo simulation.
\newblock {\em Computer methods in applied mechanics and engineering},
  191(32):3491--3507, 2002.

\bibitem{sosnovik2017neural}
Ivan Sosnovik and Ivan Oseledets.
\newblock Neural networks for topology optimization.
\newblock {\em arXiv preprint arXiv:1709.09578}, 2017.

\bibitem{khadilkar2019deep}
Aditya Khadilkar, Jun Wang, and Rahul Rai.
\newblock Deep learning--based stress prediction for bottom-up sla 3d printing
  process.
\newblock {\em The International Journal of Advanced Manufacturing Technology},
  pages 1--15, 2019.

\bibitem{solidspy}
Juan Gómez and Nicolás Guarín-Zapata.
\newblock Solidspy: 2d-finite element analysis with python, 2018.

\bibitem{masci2011stacked}
Jonathan Masci, Ueli Meier, Dan Cire{\c{s}}an, and J{\"u}rgen Schmidhuber.
\newblock Stacked convolutional auto-encoders for hierarchical feature
  extraction.
\newblock In {\em International Conference on Artificial Neural Networks},
  pages 52--59. Springer, 2011.

\bibitem{geng2015high}
Jie Geng, Jianchao Fan, Hongyu Wang, Xiaorui Ma, Baoming Li, and Fuliang Chen.
\newblock High-resolution sar image classification via deep convolutional
  autoencoders.
\newblock {\em IEEE Geoscience and Remote Sensing Letters}, 12(11):2351--2355,
  2015.

\bibitem{holden2015learning}
Daniel Holden, Jun Saito, Taku Komura, and Thomas Joyce.
\newblock Learning motion manifolds with convolutional autoencoders.
\newblock In {\em SIGGRAPH Asia 2015 Technical Briefs}, page~18. ACM, 2015.

\bibitem{24}
Yann LeCun and Yoshua Bengio.
\newblock The handbook of brain theory and neural networks.
\newblock chapter Convolutional Networks for Images, Speech, and Time Series,
  pages 255--258. MIT Press, Cambridge, MA, USA, 1998.

\bibitem{25}
Alex Krizhevsky, Ilya Sutskever, and Geoffrey~E. Hinton.
\newblock Imagenet classification with deep convolutional neural networks.
\newblock {\em Commun. ACM}, 60(6):84--90, May 2017.

\bibitem{26}
Justin Johnson, Alexandre Alahi, and Fei{-}Fei Li.
\newblock Perceptual losses for real-time style transfer and super-resolution.
\newblock {\em CoRR}, abs/1603.08155, 2016.

\bibitem{28}
Jie Hu, Li~Shen, and Gang Sun.
\newblock Squeeze-and-excitation networks.
\newblock {\em CoRR}, abs/1709.01507, 2017.

\end{thebibliography}
\end{document}